# Deep Learning, Natural Language Processing, and Explainable Artificial Intelligence in the Biomedical Domain

Milad Moradi, Matthias Samwald

{milad.moradivastegani, matthias.samwald}@meduniwien.ac.at

Institute for Artificial Intelligence
Center for Medical Statistics, Informatics, and Intelligent Systems
Medical University of Vienna, Austria

**Abstract** — In this article, we first give an introduction to artificial intelligence and its applications in biology and medicine in Section 1. Deep learning methods are then described in Section 2. We narrow down the focus of the study on textual data in Section 3, where natural language processing and its applications in the biomedical domain are described. In Section 4, we give an introduction to explainable artificial intelligence and discuss the importance of explainability of artificial intelligence systems, especially in the biomedical domain.

## 1. Artificial Intelligence

Artificial Intelligence (AI) is a scientific field that studies computer systems performing tasks that need human intelligence. AI systems aim at doing intelligent tasks such as learning from experience, knowledge representation, problem solving, reasoning, perception, and natural language processing, which are associated with cognitive functions of human mind (Russell and Norvig, 2009). Nowadays, AI systems have many applications in a wide variety of domains such as education, medicine, transportation, military, entertainment, finance, autonomous driving, smartphones, agriculture, social media, economy, healthcare, law, manufacturing, cybersecurity, and many other domains.

Since early work on artificial neurons in the 1940s (McCulloch and Pitts, 1943), different generations of AI methods have been developed. Alan Turing's theory of computation, the idea of a thinking machine, and the Turing test formed solid theoretical foundations and a milestone in early stages of development of machine intelligence in the 1950s (Turing, 1950). Then in the 1960s and 1970s, symbolic AI systems such as the General Problem Solver, the Geometry Theorem Prover, and the Logic Theorist employed high-level human-readable knowledge representation, logic programming, and reasoning to search for solutions to problems (Russell and Norvig, 2009). Knowledge-based and expert systems then became popular in the 1980s. This generation of AI systems leveraged more powerful, domain-specific knowledge (typically special-purpose rule sets in combination with customized inference methods) to solve problems in narrower fields of expertise. Meanwhile, connectionism was revived and Artificial Neural Networks (ANNs) moved the field forward towards performing more difficult and intelligent tasks, e.g. optical character recognition and speech recognition (Hopfield, 1988; Russell and Norvig, 2009). With the rapid adoption of Internet in the 1990s, AI tools such as recommender systems, search engines, and machine translation systems found their way into everyday life.

ANNs are computational methods inspired by the biological nervous systems. A neural network consists of multiple layers of interconnected processing units called "neurons" that are able to perform parallel computations. They are mostly used for data processing and knowledge representation purposes (Ramesh et al., 2004). ANNs have become attractive data analytics tools since they have the capability of handling complex, non-linear data relationships. A typical neuron in an ANN receives inputs from the previous layer in the network, aggregates the inputs, applies an activation function, and produces a final output based on the activation function's output and a threshold. Since the advent of ANNs in the 1940s (McCulloch and Pitts, 1943), artificial neurons have evolved from simple binary threshold functions to complex non-linear processing units in CNNs, RNNs, and transformer models.

In the first decades of the 21st century, availability of large volumes of data (known as "big data"), extremely powerful processors specially designed to run floating-point arithmetic computation, novel multi-layer ANN architectures, and advanced machine learning techniques have led to the development of a new class of AI methods, called "deep learning" (LeCun et al., 2015). Machine learning refers to a class of AI methods designed to learn data relationships hidden in a dataset of samples, and then generalize the learned knowledge to new unseen data samples. Accordingly, deep learning refers to a class of machine learning methods comprising multiple layers of processing units, designed to build complex generative or predictive models from large datasets. Modern deep neural networks, e.g. Convolutional Neural Networks (CNNs) and Recurrent Neural Networks (RNNs), have been widely used to model complex data relationships for a wide variety of scientific and industrial applications (Khamparia and Singh, 2019).

### 1.1. Artificial Intelligence in biomedicine and healthcare

AI systems have many applications in biomedical research and healthcare domains (Hamet and Tremblay, 2017; Min et al., 2017; Ramesh et al., 2004; Ravì et al., 2017). In many use-cases, AI methods have performed as accurate as human experts, even performed better than humans in some cases (Mintz and Brodie, 2019). ANNs are among the most widely-used AI methods utilized in the biomedical domain (Ramesh et al., 2004).

Large amounts of data in the biomedical domain make demands for developing intelligent computer methods that facilitate automatic acquisition and analysis of knowledge that is necessary to solve problems in biology and medicine. Utilizing AI systems for medical applications dates back to the 1970s, when computer analysis helped clinicians diagnose acute abdominal pain (Gunn, 1976). Since many biological, clinical, and pathological variables involve in complex interactions in biomedical datasets, ANN and deep learning models have been extensively applied to data science problems in biology and medicine. One of the first applications of ANN for clinical decision making aimed at diagnosing acute myocardial infarction (Baxt, 1990).

In clinical settings, AI systems have found many applications in diagnostic tasks such as image analysis in radiology and histopathology, waveform analysis, interpreting data in intensive care situation, and analysing Computed Tomography (CT), Magnetic Resonance Imaging (MRI), X-rays, and radioisotope scans (Amisha et al., 2019; Mintz and Brodie, 2019; Ramesh et al., 2004). AI methods have also had applications in clinical prognostic tasks such as identifying high risk patients, predicting survival chance in patients, and predicting outcome in patients with specific cancers (Ramesh et al., 2004). Another type of AI systems, i.e. AI-robots, have been used as assistant-surgeons or solo performers in surgery, or as care-bots for assisting in the delivery of care (Amisha et al., 2019; Hamet and Tremblay, 2017). They can be also used to monitor the guided delivery of drugs to target tumours, tissues, or organs (Hamet and Tremblay, 2017). In biomedical research, machine learning methods have had applications in discovering novel therapeutic targets using protein-protein interaction algorithms (Theofilatos et al., 2015), identifying DNA variants as predictors of a specific disease (Rapakoulia et al., 2014),

Image and text are two data modalities in the biomedical domain that are commonly analysed using machine learning methods. AI-driven image processing have many applications in radiology, oncology, cardiology, gastroenterology, ophthalmology, etc. (Mintz and Brodie, 2019). Applications of AI systems in biomedical and clinical text processing is extensively discussed in Section 3.3.

## 2. Deep learning

Deep learning refers to a subclass of machine learning methods that use multiple layers of processing units, usually known as neurons, motivated by the working of animal neocortex (Yamins and DiCarlo, 2016). The goal of a deep learning model is to learn a hierarchy of concepts such that in every layer of the model a continuous vector representation encodes a different level of abstraction (LeCun et al., 2015). Traditional machine learning algorithms were not able to process natural data, e.g. image and language, in their raw form. For decades, large amounts of engineering efforts and domain knowledge were needed to design feature extractors that can transform raw data into proper feature vectors or internal representations that can be processed by machine learning and pattern recognition algorithms (LeCun et al., 2015). Nowadays, thanks to advanced deep learning algorithms, machine learning systems are able to automatically learn feature vectors and internal representations that can be used to detect and classify intricate non-linear relationships in a dataset. Processing units in different layers of a deep neural network learn to represent the network's inputs in a way that enables the model to correctly predict the target output. Various neural network architectures are used to implement deep learning algorithms for different applications (Khamparia and Singh, 2019). Some of the most widely-used deep neural network architectures are described in the following.

A **deep feed-forward neural network** (or **multi-layer perceptron**) has the simplest architecture of a deep learning model. It is made up of an input layer, multiple hidden layers, and an output layer. Nodes in the input layer perform no computation and just pass on the input data to the hidden nodes. Hidden layers are responsible to do non-linear transformations on the input data. The output layer finally estimates output probabilities for possible outcomes. The network is initialized with random weights, and the gradient descent algorithm is used to train the network and find an optimal set of weights. Figure 1 illustrates the schematic diagram of a fully-connected deep feed-forward neural network. Training of deep neural networks is generally performed in three steps:

1) In the forward pass, the data is feed to the network. Starting from the input layer, training samples pass through hidden layers, and an output probability is finally estimated by the output layer.
2) A loss function computes the error by comparing the network's output and the target output.
3) In the backward pass, the weights (or parameters) of the network are adjusted in the opposite direction of the gradient vector, which specifies, for every weight, the amount of increase or decrease in the network's error if the weight were increased by a small amount.

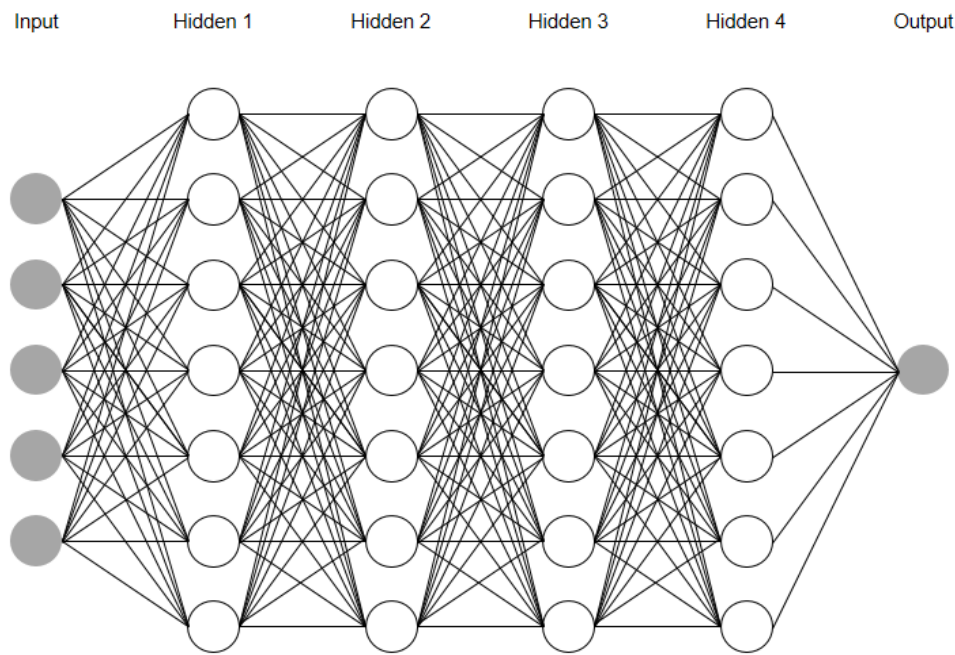

Figure 1: Schematic diagram of a fully-connected deep feed-forward neural network with four intermediate hidden layers. Each node in a layer is connected to all nodes in the previous layer.

A **Restricted Boltzmann Machine** (**RBM**) is a variation of neural networks with stochastic processing units that are connected bidirectionally. RBMs are generative stochastic neural networks that are able to learn a probability distribution over their inputs. There are two groups of neurons in a RMB, i.e. visible and hidden neurons. These neurons must form a bipartite graph in which there are symmetric connections between visible and hidden units, but there are no connection between nodes within a group of neurons. This property makes RBMs suitable for more efficient training algorithms than those that are used for training unrestricted Boltzmann machines. Training of a RBM starts with a random state in one layer, then the Gibbs sampler is used to generate data from the RBM. Given the states of units in one layer, all units in other layers are updated in an iterative manner until the equilibrium distribution is reached. Finally, the weights within the RBM are computed by maximizing the likelihood of the RBM. Figure 2 shows the schematic diagram of a RMB with two groups of visible and hidden units. A deep RBM can be built by stacking multiple RBMs on top of each other. RBMs are generally used for feature extraction in classification tasks. They have also found applications in dimensionality reduction, collaborative filtering, and topic modelling (Liu et al., 2017).

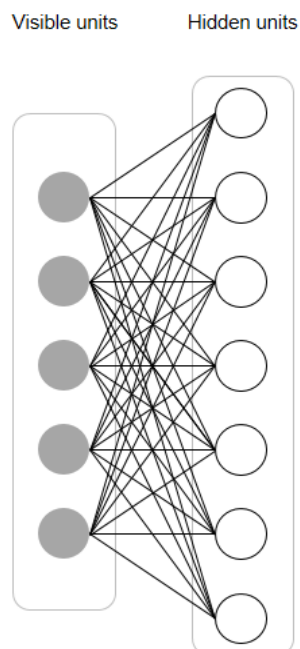

Figure 2: Schematic diagram of a Restricted Boltzmann Machine (RBM). There are symmetric connections between visible and hidden units, but there is no connection between units within the same layer.

A **Deep Belief Network** (**DBN**) is a probabilistic generative graphical model that consists of multiple layers of hidden units (latent variables) such that there are connections between layers but not between units within each layer. DBNs are special forms of RBMs, however, there are some differences. The main difference is that hidden and visible variables are not mutually independent in a RBM, hence, the network is undirected. On the other hand, a DBN allows exploring dependencies between variables by having a directed topology. Both RBMs and DBNs are trained using layer-wise greedy training. DBNs are usually trained in an unsupervised manner that allows probabilistic reconstruction of its inputs. After training, the layers can act as feature detectors, however, they can be also further trained on labelled data to perform classification tasks.

Figure 3 shows the schematic diagram of a DBN. It can be seen from Figure 3 that a DBN is a variation of stacked RBMs with undirected connections between the top two layers, and directed connections between lower layers.

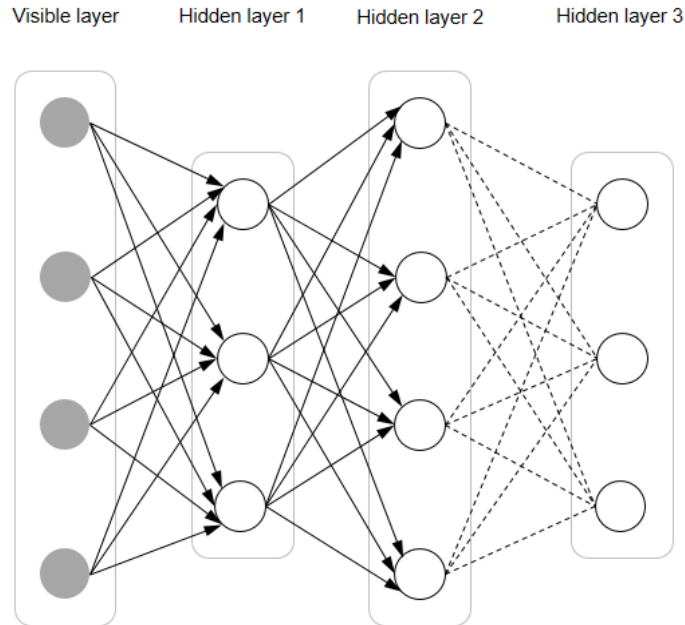

Figure 3: Schematic diagram of a Deep Belief Network (DBN) with one visible layer and three hidden layers. There are undirected connections (represented by dash lines) between the top two layers. Connections between the lower layers are directed.

An **autoencoder** is an unsupervised neural network that is used for dimensionality reduction and data compression. Autoencoders are able to learn generative models of data. An autoencoder converts the input data into an abstract representation (also known as code) using recognition weights in the encoding phase. It then reconstructs the input through converting back the abstract representation into the original data using a generative set of weights in the decoding phase. Autoencoders and multi-layer perceptrons have similar architecture; they also use similar training and propagation mechanism. However, the goal of an autoencoder is to reconstruct the input, but the aim of a multi-layer perceptron is to predict the target value with respect to the input (Khamparia and Singh, 2019; Liu et al., 2017). The training process of an autoencoder can be performed in two steps: 1) to learn features through unsupervised training, and 2) to fine-tune the network through supervised training. Figure 4 shows the schematic structure of a deep autoencoder neural network.

A **Convolutional Neural Network** (**CNN**) is a variant of artificial neural networks inspired by the organization of the animal visual cortex. CNNs have shown to be ideal models for processing two-dimensional data with grid-like topology, e.g. images and videos (Liu et al., 2017). There are two different types of layers in a CNN, i.e. convolution and subsampling layers. Figure 5 illustrates the schematic structure of a CNN. The first convolution layer convolves the input image with trainable filters at all possible offsets, then local feature maps are produced. The feature maps are then subsampled using a pooling mechanism, and the size of feature maps reduces in the next subsampling layer. Higher level transformations of the input are constructed as feature maps pass through consecutive convolution and subsampling layers. Finally, the values of the input pixels are rasterized and represented in the form of a single vector, which is used as the input of a fully-connected feed-forward network to perform classification or other tasks. CNNs are trained using backpropagation and gradient descent, similar to standard neural networks.

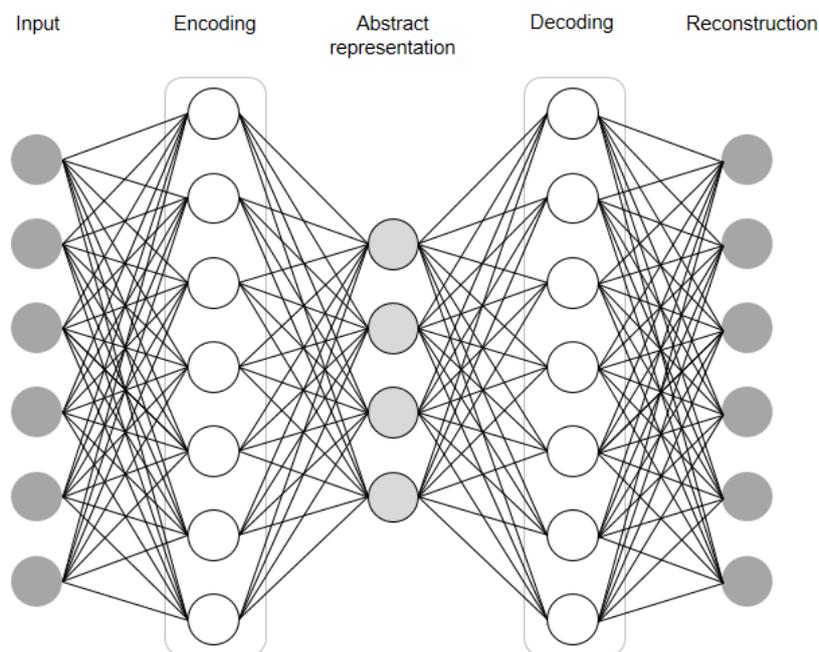

Figure 4: Schematic diagram of a deep autoencoder neural network.

A **Recurrent Neural Network** (**RNN**) is able to process sequential data, such as texts and time series, where it is important to capture sequential characteristics of data and use temporal patterns to predict the next likely scenario. The schematic structure of a RNN and the unfolding in time are illustrated in Figure 6. As can be seen, in every layer, there is a state vector that implicitly stores the context seen by the processing unit until that time step. The state vector is passed to the same processing unit in the next time step, then, with respect to the input at that time step, the state vector is updated and an output is produced that will be used as the input of the next layer. Since RNNs can process samples with different lengths of input and output, they have been shown to be more appropriate than standard feed-forward neural networks for NLP applications. The basic RNN model cannot perfectly capture long-term dependencies in long input texts due to the vanishing gradient problem. **Gated Recurrent Unit** (**GRU**) and **Long Short-Term Memory** (**LSTM**) networks have been able to effectively address the vanishing gradient problem by utilizing memory cells and gating mechanisms that control how long previous knowledge should be stored for being used to process future words, and when previous knowledge should be updated or forgotten. A GRU is a simplified version of a LSTM, so it needs less computation. Therefore, GRUs are more scalable than LSTMs. However, LSTMs are more powerful and flexible than GRUs.

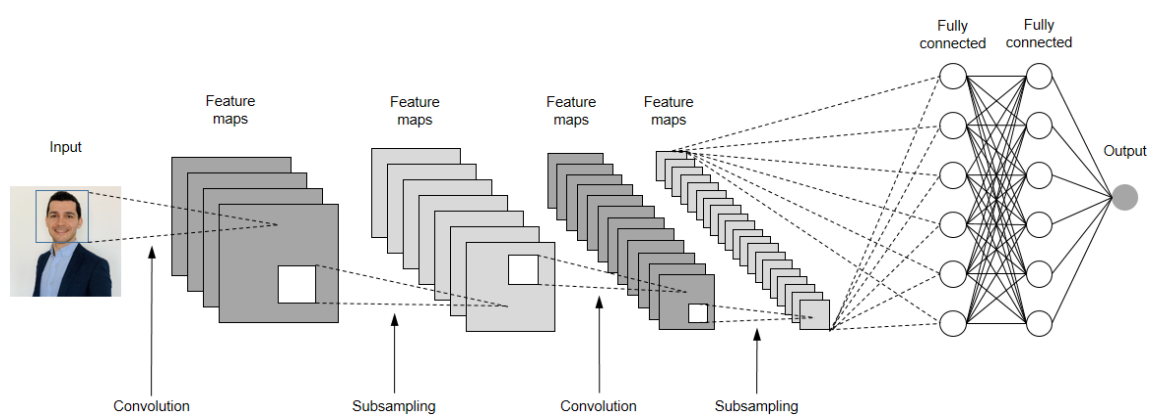

Figure 5: Schematic structure of a Convolutional Neural Network (CNN).

Deep learning methods have shown massive capability in discovering complex structures in high-dimensional data, hence, they have found many applications in various fields of science, education, business, medicine, transportation, and other domains. Deep learning techniques have significantly improved AI systems for several biomedical applications such as, predicting the effects of mutations in non-coding DNA on gene expression and disease, predicting the activity of potential drug molecules, reconstructing brain circuits, protein structure prediction, gene expression regulation, biomedical imaging, biomedical signal processing, translational bioinformatics, cancer diagnosis, cell clustering, pervasive sensing, modelling infectious disease epidemics, medical information extraction, and many other applications (Hahn and Oleynik, 2020; LeCun et al., 2015; Liu et al., 2017; Min et al., 2017; Ravì et al., 2017).

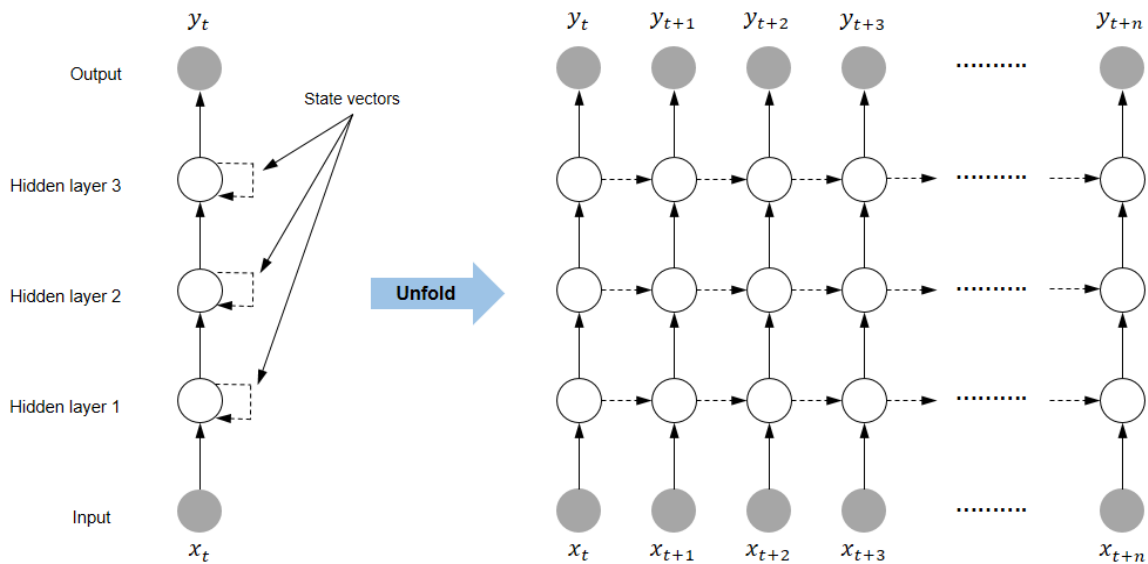

Figure 6: Schematic diagram of a deep Recurrent Neural Network (RNN) and its unfolding in time.

## 3. Natural Language Processing

Natural Language Processing (NLP) lies in the intersection between linguistics and artificial intelligence, and refers to designing and developing computerized methods that are able to process, analyse, understand, and generate human language. NLP provides a set of computational techniques to facilitate automatic analysis and representation of human language ([Chowdhary, 2020](#)).

Since the earliest works on NLP in the early 1950s until the late 1980s, NLP systems relied on symbolic methods. These methods utilized complex sets of hand-crafted rules, sometimes in combination with conceptual ontologies, to process textual data. Starting from the early 1990s, with the rapid adoption of machine learning, NLP was revolutionized and statistical methods became widespread for developing automatic text processing systems. Machine learning methods helped NLP scientists develop language processing systems that automatically learn rules through applying statistical inference to large volumes of text data. However, rule-based methods were not completely abandoned and were still used for preprocessing, e.g. tokenization, or for postprocessing, e.g. knowledge extraction, in a NLP pipeline.

Since the early 2010s, thanks to development of powerful deep learning and representation learning methods, neural NLP systems have become popular and been widely utilized for various language processing applications. Deep learning models remove the demanding task of feature engineering that was a big challenge of building statistical NLP models. Word embeddings learned by deep language models can encode semantic and syntactic properties of words. They can be then used as the input of any NLP systems, regardless of the task at hand. Nowadays, neural NLP models are used in a wide variety of applications such as information retrieval, named entity recognition, word sense disambiguation, information extraction, relation extraction, natural language translation (or machine translation), question answering, summarization, sentiment analysis (or opinion mining), topic modelling, text classification, dialogue systems, and speech recognition (Chowdhury, 2003; Nadkarni et al., 2011). There are various low-level subtasks in NLP that are likely to be found as parts of a high-level task in any text processing systems. These subtasks include negation detection, temporality detection, context determination, hedging, sentence boundary detection, word tokenization, lemmatization, part-of-speech tagging, stemming, parsing, semantic role labelling, and coreference resolution.

## 3.1. Deep learning in natural language processing

As discussed in Section 2, RNN and its variants have been widely used for implementing NLP systems, since they can effectively handle long-term dependencies in text sequences. The generic RNN illustrated in Figure 6 is able to pass forward the context it has seen until the current time step through forward connections. This allows the network to be aware of the context when it processes next words in the input sequence. However, it is sometimes beneficial to process the current word with respect to the context that follows. The idea of having both forward and backward connections has led to a more powerful variant of RNNs, called **bidirectional RNNs** (**bi-RNNs**). Figure 7 shows the schematic diagram of deep bi-RNN with forward and backward layers. The output of the last forward and backward layers are combined using a fully-connected layer, and the output is generated for every time step. The generic bi-RNN model presented in Figure 7 can be generalized to bi-LSTM and bi-GRU models. Embeddings from Language Models (ELMo) is a very large NLP model composed of many forward and backward LSTM layers that set a breakthrough in language processing by providing contextualized word vector representations (Peters et al., 2018).

For more complex NLP tasks such as machine translation, summarization, or captioning that need to generate a new textual output, **encoder-decoder** deep neural networks are the most widely-used models (Otter et al., 2021). An encoder-decoder sequence-to-sequence (seq2seq) model is composed of two main parts, i.e. encoder and decoder. The encoder converts the input sequence into a numeric representation that is fed into the decoder whose job is to generate a new text with respect to the encoded input. RNN variants are commonly used to implement encoders and decoders. Figure 8 illustrates the schematic diagram of an encoder-decoder neural network composed of an encoder and a decoder RNN model. In this example, a bi-RNN is used for the encoder and a forward RNN is used for the decoder. As can be seen, the output of the decoder at each time step is used as the input of the decoder at the next time step.

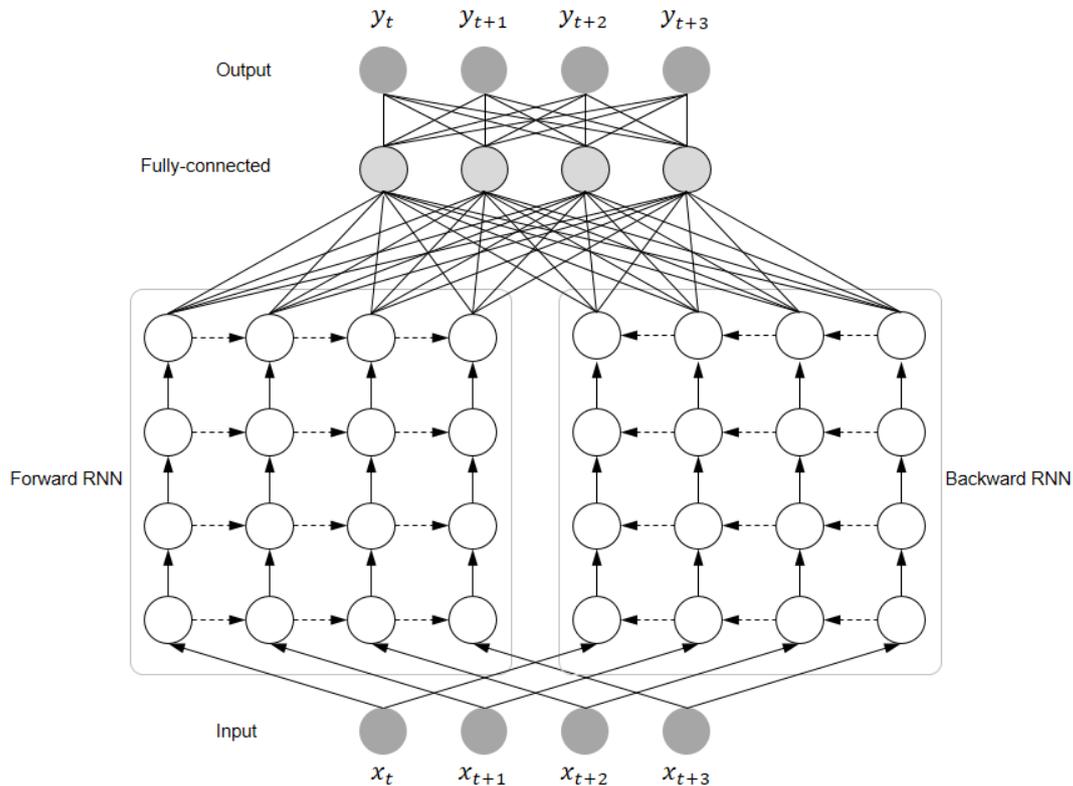

Figure 7: Schematic diagram of deep bidirectional Recurrent Neural Network (bi-RNN).

As already discussed, RNN and its variants are able to capture long-term dependencies in sequential data, which has advanced NLP systems by enabling them efficiently process textual data and achieve higher levels of performance on complicated tasks. However, for some applications it is required to specify if any of the inputs are more important than others when the next output is produced. **Attention mechanism** can properly address this problem by allowing the neural network to learn which part of the input is more important than others with respect to the context and the task at hand (Vaswani et al., 2017). The attention mechanism can be added to a RNN network as an additional layer that has access to state vectors of all previous inputs and assigns weights to each previous state. These attention weights indicate which parts of the input are more relevant and should have more contribution to computing a state vector and output for the next input. So far, different variants of attention mechanism such as self-attention, cross-attention, multi-heads, query-key-value, and dot-products have been implemented for various applications.

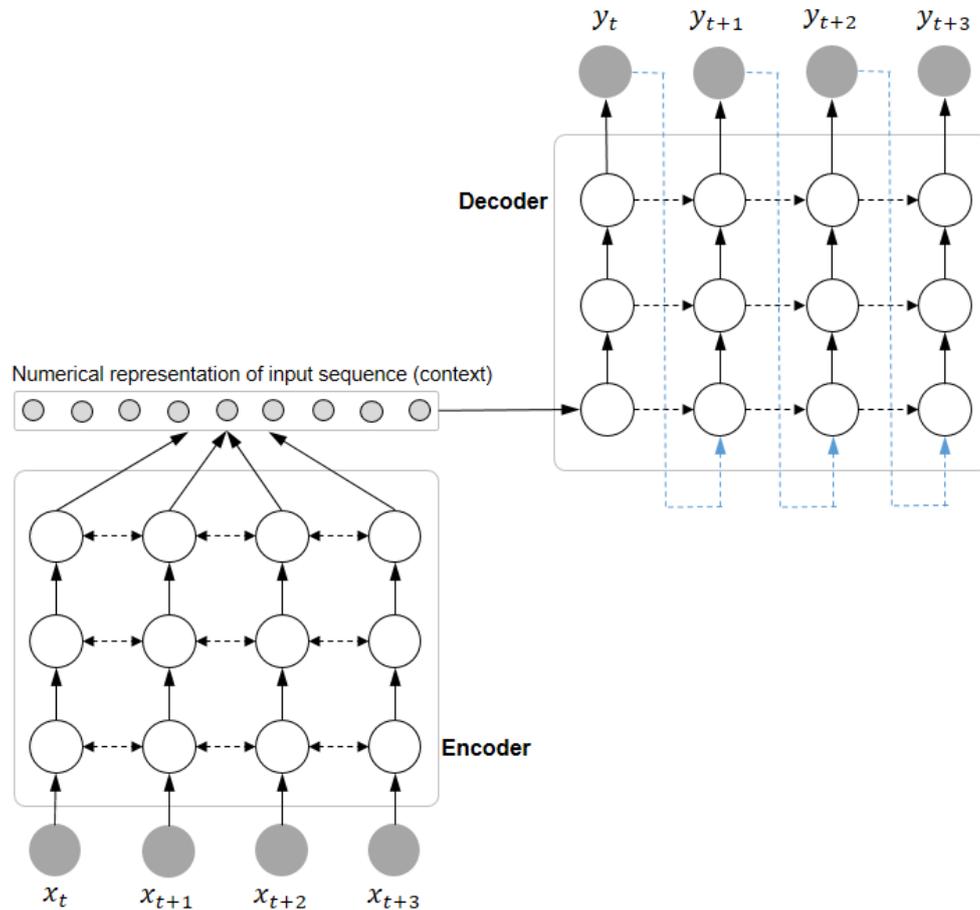

Figure 8: Schematic diagram of a deep encoder-decoder neural network composed of an encoder and a decoder RNN model.

A major disadvantage of RNNs is that they need to process the input sequence in order because token computations depend on results of previous token computations. This is a barrier to parallelization of computations, which makes the training of RNNs inefficient. In the recent years, **transformer** neural networks have effectively addressed the weaknesses of RNNs by adopting self-attention and removing RNN structure (Vaswani et al., 2017). In contrast to RNN, a transformer does not need to process the input sequence in order. Since the self-attention mechanism provides context for any word in the input, the transformer model can benefit from parallelization. This additional parallelization has enabled training very large language models on massive training corpora. BERT (Bidirectional Encoder Representations from Transformers) (Devlin et al., 2018) and GPT (Generative Pre-trained Transformer) (Radford et al., 2019) are two large language models that utilize transformers. They have achieved state-of-the-art accuracy scores on many NLP tasks.

Figure 9 shows the schematic diagram of a transformer encoder-decoder neural network. In this example, there are three encoder layers, each one composed of a self-attention and a feed-forward layer. The self-attention layer receives input encodings from previous encoder layer and weighs the relevance of inputs to each other. The output encoding is then further processed by a feed-forward layer. The encoder uses positional encoding because it needs to know about the order of the sequence. The decoder has three decoder layers, each one composed of a self-attention, an encoding-attention (cross-attention), and a feed-forward layer. The encoding-attention layer draws relevant information from the encoding computed by the encoder. The first decoder layer needs positional encoding and embeddings of the output sequence.

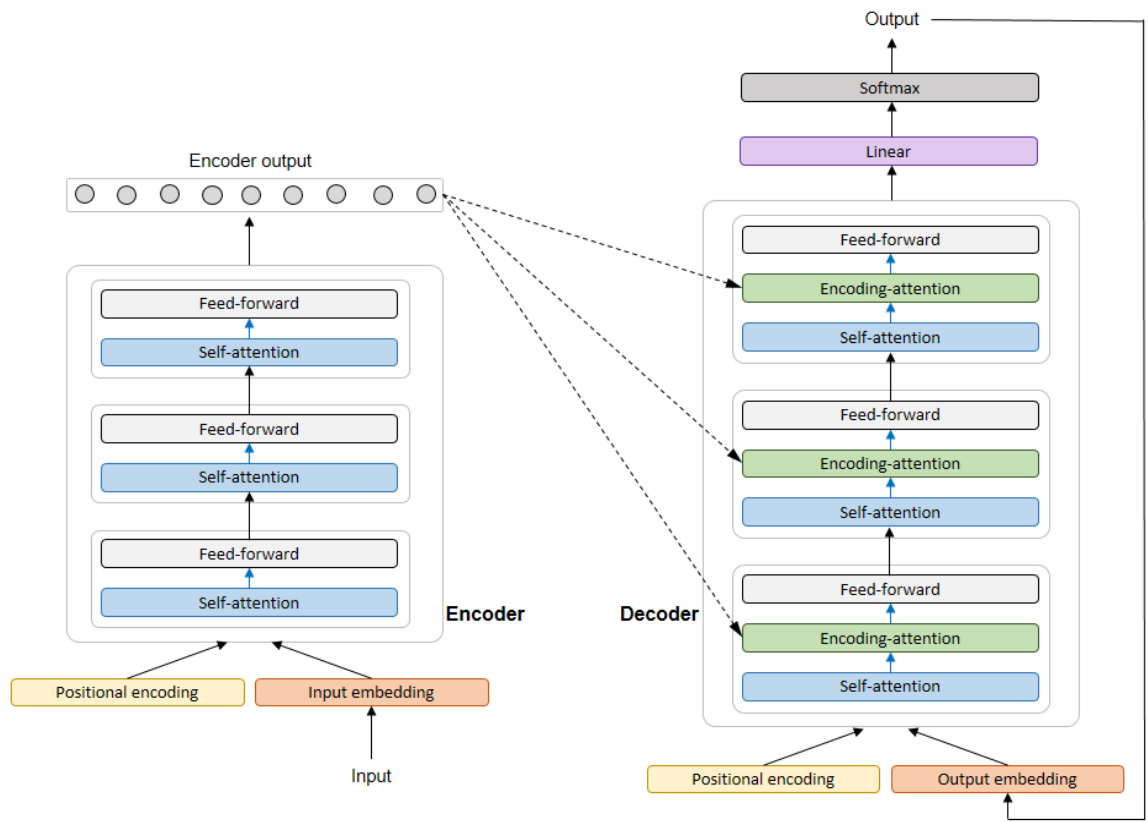

Figure 9: Schematic diagram of a transformer encoder-decoder neural network.

Deep learning neural networks explained in this section have been widely used to implement powerful NLP systems for a wide variety of applications such as Information retrieval, named entity recognition, information extraction, text classification, text generation, summarization, question answering, and machine translation (Otter et al., 2021). One major application of deep learning in NLP is language modelling. A language model is able to predict words or linguistic properties given previous words or linguistic information. A language model learns to encode lexical, syntactic, and semantic information by being trained on large text corpora. The learned knowledge is encoded in the form of word vector representations, namely embeddings (Mikolov et al., 2013).

*3.2. Vector representation of text*

Before the introduction of deep neural language models, statistical NLP models used features that relied on counting how many times sequences of words or characters of length up to N (known as N-grams) occur in a corpus of text documents (LeCun et al., 2015). Given a vocabulary with size V, the number of possible N-grams is on the order of $V^N$, therefore, very large text corpora would be needed to train a language model that is able to handle a context of more than a handful of words (LeCun et al., 2015). A major problem of N-grams is that they treat each word as an atomic unit, hence, they are not able to identify semantically related sequences of words. On the other hand, neural language models are able to generalize across semantically related word sequences because they associate each word with a continuous vector representation such that semantically related words are close to each other in the vector space.

Much attention has been devoted to word representation learning since it is one of the main research areas in NLP. Early vector space models of word representation dealt with documents as vectors whose dimensions were the whole vocabulary, and the value of each dimension was computed as some sorts of the frequency or normalized frequency of the respective word within the document. Document-based vector space models were also extended to smaller units such as words. A word-based vector space model is usually constructed based on the normalized frequencies of co-occurring words in a corpus (Camacho-Collados and Pilehvar, 2018). The main goal of building a word-based vector space model is to represent words as numerical vectors such that those words that share similar context should be close in the vector space.

A main disadvantage of traditional word representations is the high dimensionality. The number of dimensions in a word vector can easily reach hundreds of thousands, or even millions, because dimensionality corresponds to the number of words in the vocabulary. However, a solution is to perform dimensionality reduction using Singular Value Decomposition (SVD) and Latent Semantic Allocation (LSA). Neural networks have been utilised to enable learning low-dimensional word vectors through unsupervised learning. Word2Vec was one of the first methods that exploited neural networks to construct low-dimensional and efficient vectors for meaning representation (Mikolov et al., 2013). Word embeddings learned by Word2Vec can be then used as the input of a down-stream neural network performing a supervised NLP task, e.g. classification or sentiment analysis. GloVe is another major word embedding method that employs a bilinear regression model in combination with global matrix factorization and local context window methods (Pennington et al., 2014).

In spite of achieving high performance scores on several NLP tasks, the above word embedding methods has a major limitation: they ignore the fact that a word can have multiple meanings and the meaning can change based on the context. These models generate a single vector representation for each word, regardless of its context. Contextualized word embeddings address this problem by enabling dynamic word vectors that change depending on the context in which they appear (Camacho-Collados and Pilehvar, 2018). ELMo (Peters et al., 2018) and BERT (Devlin et al., 2018) set major breakthroughs in NLP by providing novel contextualized vector representation methods. The former utilizes backward and forward LSTMs, and the latter uses transformers and attention mechanism to capture context.

In recent years, context-free and context-sensitive word embeddings have been trained on biomedical and clinical text corpora, in order to be used as input in down-stream biomedical or clinical NLP tasks (Khattak et al., 2019; Tawfik and Spruit, 2020; Wang et al., 2018a). BioBERT (J. Lee et al., 2019b), SciBERT (Beltagy et al., 2019), and BioELMo (Jin et al., 2019) are among language models that were pretrained on biomedical text and improved the performance of several biomedical NLP tasks. Moreover, ClinicalBERT (Alsentzer et al., 2019), ClinicalXLNet (K. Huang et al., 2020), and ClinicalELMo (Peters et al., 2018) were pretrained on clinical notes to be used for clinical text analysis tasks.

### 3.3. Natural language processing in the biomedical domain

In the biomedical domain, textbooks and published peer-reviewed scientific papers are valuable sources of knowledge (Fleuren and Alkema, 2015; Mintz and Brodie, 2019). Clinical notes, electronic health records, lab reports, and experimental data are other sources of information that need to be stored in digital format to be subsequently processed and analysed in biomedical research (Fleuren and Alkema, 2015). With the extremely large volumes of textual documents that are available in the biomedical domain, it is almost impossible to manually retrieve relevant information and analyse it in a timely manner. In order to address this challenge, numerous automatic biomedical text processing systems have been developed using NLP and text mining methods.

NLP and text mining methods in the biomedical domain are divided into two main categories, i.e. 1) rule-based or knowledge-based, and 2) statistical or machine learning based methods (Cohen, 2014). Knowledge-based biomedical NLP systems mostly use two types of knowledge to solve a NLP problem: 1) knowledge about the language, it defines how facts are stated in biomedical texts, and 2) real-world knowledge about the domain that is defined in the form of rule sets or ontologies (Cohen, 2014). Rules usually define patterns of properties that need to be fulfilled by some text in a document (Wang et al., 2018b). Machine learning based methods, on the other hand, automatically learn domain and language knowledge from a set of labelled training examples and generalize that knowledge to new unseen examples. Biomedical NLP systems can also utilize a combination of knowledge-based and machine learning approaches, these systems are called hybrid systems. A main disadvantage of knowledge-based methods is that considerable manual effort is needed to build numerous rules that form a knowledge base. In contrast, little or no domain-specific knowledge is needed to build a machine learning based biomedical NLP system. They learn the underlying knowledge in the data by discovering statistically tractable phenomena and relations between them.

Some applications of NLP and text mining methods in the biomedical and healthcare domains include finding protein-protein interactions, detecting disease-treatment relations, summarizing biomedical literature or clinical documents, cohort retrieval and phenotype definition, genome and gene expression annotation, drug-target discovery, drug repositioning, exploring possible adverse events of drugs, and discovering drug side-effects and interactions by analysing electronic health records (Cohen, 2014; Fleuren and Alkema, 2015; Wang et al., 2018b).

Many NLP and text mining applications are built on MEDLINE since it provides a rich programming interface and annotated abstracts with Medical Subject Heading (MeSH) terms (Fleuren and Alkema, 2015). Medline Ranker (Fontaine et al., 2009), askMEDLINE (Fontelo et al., 2005), PICO (K.-C. Huang et al., 2013), PubCrawler (Hokamp and Wolfe, 2004), PubFocus (Plikus et al., 2006), ProteinCorral (Li et al., 2013), PubViz (Xuan et al.), CoPub (Fleuren et al., 2011), and PPInterFinder (Raja et al., 2013) are some examples of NLP and text mining tools built on MEDLINE to do various tasks such as querying MEDLINE/PubMed for retrieving and ranking relevant abstracts, scanning daily updates and alerting users about new relevant items in the database, enriching user's queries by additional information from domain knowledge sources, identifying protein names within abstracts, search result visualization by connecting queries to concepts in a knowledge base, extracting gene lists from abstracts, and finding protein-protein interactions within abstracts.

Clinical documents appear in different types, e.g. discharge summaries, ICU progress notes, pathology reports, X-ray reports, and Electronic Health Records (EHRs) (Cohen, 2014). Applying NLP and text mining methods to various types of clinical textual data have had a wide range of applications in clinical decision support, quality improvement, and clinical/translational research (Wang et al., 2018b). During the last decades, NLP methods have been successfully employed for cancer-case identification, staging, and outcomes quantification (Yim et al., 2016), identifying patient phenotype cohorts (Kumar et al., 2014), temporal relation discovery (Lin et al., 2016), detecting medication discrepancies (Feng et al., 2015), risk factor identification (Khalifa and Meystre, 2015), pharmacoepidemiology (Salmasian et al., 2013), symptom identification (Dreisbach et al., 2019), and clinical workflow optimization (Popejoy et al., 2014).

There are also several shared tasks designed to advance NLP methods for clinical applications. These shared tasks address various applications, such as clinical named entity recognition (Pradhan et al., 2013), automatic de-identification of personal health information (Uzuner et al., 2007), identification of obesity and its co-morbidities (Uzuner, 2009), identification of medications and their dosages (Uzuner et al., 2010), clinical concept extraction and relation classification (Uzuner et al., 2011), automatic identification of medical risk factors (Stubbs et al., 2015), and temporal information extraction from clinical texts (Bethard et al., 2016).

In the next subsection, we discuss biomedical text summarization that is a major task in biomedical NLP.

### 3.4. Biomedical text summarization

Recent years have witnessed an exponential growth in the amount of information that is available to clinicians and biomedical researchers. The biomedical literature and EHRs are two main sources of textual information in the biomedical domain. Retrieving, interpreting, and integrating relevant information from large volumes of text documents is almost impossible without employing automatic tools. Automatic text summarization tools can help reduce time and effort required to identify and process important information from lengthy text documents. A summary is a transformation of the source text into a shorter text that conveys the most important/relevant content of the source (Mishra et al., 2014).

Text summarization methods are generally divided into two categories, i.e. extractive versus abstractive methods. An extractive summarizer identifies the most important sentences in one or multiple input text documents, extracts and concatenates them to form the final summary (Moradi et al., 2020a). In contrast, an abstractive summarizer generates a summary by producing new sentences that are different to the original sentences but convey the most important points (El-Kassas et al., 2021). Regarding the number of documents that a summarization system receives as the input, a summary can be produced for either a single document or multiple documents. Multi-document summarization can be more challenging than single-document summarization, since repetitive information may appear in several documents. Therefore, the summarizer should be capable of handling information redundancy. Another challenge is that information should appear in the summary with respect to the temporal order in which information appeared in the original documents. Cohesion and coherency are other challenges that need to be addressed when producing a unified summary for multiple documents (El-Kassas et al., 2021; Mishra et al., 2014; Moradi, 2018).

A summary can be either query-based or generic. A query-based summary presents content that is most relevant to a query or a set of keywords given by user. On the other hand, a generic summary presents a general sense of the most important content of the original text (El-Kassas et al., 2021). Summarization systems can be built in either supervised or unsupervised manner. Annotated training data are needed to train a summarizer in the supervised scenario, while unsupervised summarization does not need to be trained on training data (El-Kassas et al., 2021). The former is more demanding and expensive because it needs a lot of human efforts for manual annotation. Summarization systems can be divided into general versus domain-specific groups. General summarizers are designed to summarize documents from different domains, while domain-specific summarizers are specialized to summarize documents that belong to a certain domain, e.g. biomedical documents (El-Kassas et al., 2021).

Initial work in text summarization relied on simple term frequency features to identify important sentences within a document. Since then, various statistical features and heuristics have been employed for content selection in text summarization. Position of a sentence in a document, length of a sentence, presence of predefined cue phrases in a sentence, keywords extracted from input document, presence of title words, centroid-based cohesion, and co-occurrence feature are among widely-adopted features and heuristics for text summarization (Moradi et al., 2020b). It has been shown that these general features may not properly identify important contents in biomedical text summarization (Moradi and Ghadiri, 2017, 2018; Plaza et al., 2011; Reeve et al., 2007). In order to address this issue, biomedical text summarizers employed sources of domain knowledge to map input text to a concept-based representation (Ji et al., 2017; Mishra et al., 2014; Moradi and Ghadiri, 2018; Plaza et al., 2011; Zhang et al., 2011). This helps assess the informative content of text with respect to the context and semantics of words/sentences, rather than shallow features used by traditional summarizers.

However, concept-based biomedical text summarization that utilizes sources of domain knowledge have their own issues. Building, maintaining, and utilizing knowledge bases may be time-consuming and expensive. Large amounts of human effort are needed to define biomedical concepts and relations between them. Moreover, the choice of appropriate source of domain knowledge is a challenge that may have a significant effect on the performance of a biomedical text summarization system (Plaza, 2014). Another challenge is how to measure the importance of words and sentences based on qualitative relations between concepts (Moradi and Ghadiri, 2017). Deep learning and representation learning models have addressed these challenges by representing textual data as continuous representations such that words are encoded into numerical vectors based on the context in which they appear (Fathi and Maleki Shoja, 2018; Otter et al., 2021). In this way, the similarity, relatedness, and importance of words and sentences can be quantified with respect to their semantics and the context.

## 4. Explainable artificial intelligence

As already discussed in this article, deep neural networks have outperformed other classes of machine learning methods for a wide variety of tasks. They have performed even better than humans on some specific tasks, especially for image and text processing. However, one main challenge of deep learning models is that they are black-box. In other words, they hide their internal logic to the user, and their internals are uninterpretable by humans (Guidotti et al., 2018). Although the underlying mathematical principles of deep learning models are known, it is difficult to represent the knowledge learned by such models or to explain their decision making process (Holzinger et al., 2017). The increasing complexity of predictive models has led to some difficulties when explaining and interpreting the behaviour of these models (Lakkaraju et al., 2017). Deep learning models, or black-box models in general, have a great ability to learn complex patterns that exist in the data, then predict new unobserved instances. Their predictive power can be complemented by the ability to explain or interpret what they have learned (Murdoch et al., 2019). Consequently, it may be easier to trust them or optimize their performance. Users of AI systems need to 1) decide when and how they should trust a predictive model, 2) detect potential biases in the model and data, and 3) take actions to refine the model. Therefore, they should have a clear understanding of the model and its behaviour (Lakkaraju et al., 2017). Moreover, predictive machine learning models are usually trained on data that may contain biases. Consequently, models learned on such data inherits the biases, leading to wrong or unfair decisions (Guidotti et al., 2018).

Interpretability can be defined as the ability to present or to explain the behaviour, rational, decision making process, or internal working of a machine learning model in a way that can be understood by a human (Doshi-Velez and Kim, 2017). Interpretable machine learning can be also used for extracting relevant knowledge about domain relationships conveyed by the data (Murdoch et al., 2019). Explanations can be represented as mathematical equations, visualizations, heatmaps, saliency masks, partial dependency plots, decision trees, rule sets, feature importance scores, or natural language.

There are various reasons to develop AI systems whose decisions are transparent, understandable, and explainable:

*1) Trust*

It is needed to understand reason(s) why a decision was made by an AI system in order to trust the system and its outputs. We need to ensure that the AI system makes correct decisions for the right reasons.

*2) Ethics and fairness*

Explainable AI (XAI) can be used to make sure decisions made by an AI system are fair and ethical. It can also help to ensure that critical data, such as race, gender, etc. do not affect algorithmic decisions in undesired ways.

*3) System performance*

Being aware of potential weaknesses of an AI system, it is easier to improve its performance. When it is known what the system is doing, why it sometimes fails, and why it makes its decisions, it will be easier to detect biases in the model or in the data (Moradi and Samwald, 2021a). This also helps to have a better understanding of the problem at hand.

*4) Safety and security*

Organizations need to make sure only permitted data are used for agreed purposes. XAI helps to make sure the AI system complies with certain intended requirements. Using XAI, organizations can have control over their AI systems. Hacking attacks,

Deep Learning, Natural Language Processing, and Explainable Artificial Intelligence in the Biomedical Domainengineering oversights, and deliberate unethical design can be identified using explainability methods that clearly show the decision making process and factors affecting the output of the AI system.

*5) Accountability*

This requirement concerns who is accountable for decisions made by an AI system. Using XAI, it is possible to trace a chain of causality from the AI system back to the person or organization. This can help to assign responsibility for the decisions made by the AI system.

*4.1. Model-based versus post-hoc explainability*

Explainability by design, or model-based explainability, refers to incorporating transparency into the internal working and decision making process of a model. Explainability or interpretability considerations in this step can mostly concern the choice between simple, easy to interpret models and complex, black-box models that fit the data more accurately (Murdoch et al., 2019). Linear regression models, decision sets, decision trees, and generalized additive models are typical examples of transparent or glass-box AI models. Model-based explainability considerations may impose the model designer to put limitations on the complexity of the AI model, subsequently the predictive accuracy my decrease. Therefore, model-based interpretability should not be considered when the underlying relationship is complex, because a simple model cannot fit complex data well (Murdoch et al., 2019). Two criteria often play key roles in selecting a model in this step: 1) **simplicity**, in order to be human-understandable, and 2) **complexity**, in order to fit the underlying data relationships.

Model-based interpretability can be further divided into more specific levels, i.e. simulatability, decomposability, and algorithmic transparency (Lipton, 2016). **Simulatability** concerns how transparent and human understandable the entire model is. **Decomposability** refers to the explainability of individual components of a model, like inputs, parameters, and calculations. **Algorithmic transparency** is applied at the level of the learning algorithm. As an example, it is provable that a linear model will converge to a unique solution even for new problems and datasets. On the other hand, the heuristic optimization procedures used in training neural networks cannot guarantee that they will find unique solutions for new problems (Lipton, 2016). Full transparency may be feasible only for small and simple models, e.g. shallow decision trees, simple linear models, or rule bases (Preece et al., 2018).

Post-hoc explainability refers to providing explanation for a specific decision, or a set of decisions, made by the model. This allows the reproduction of decisions when needed (Holzinger et al., 2017). In post-hoc interpretability, there is a trained model as input, and the goal is to extract information about what relationships the model has learned (Moradi and Samwald, 2021c; Murdoch et al., 2019). Post-hoc explainability can be most useful when there are complex relationships in the data and we need to build an intricate, black-box model. Using post-hoc explanations we do not sacrifice the predictive accuracy of the model because it is not required to make explainability considerations during the modelling phase (Lipton, 2016).

Post-hoc explanation methods can be divided into two categories, i.e. global and local methods (Murdoch et al., 2019):

1) **Global** or **dataset-level** interpretation refers to approximating the global behaviour of a predictive model. These methods are more helpful when we are interested in general relationships learned by the model, e.g. what features or interactions are associated with a particular class in the whole dataset or in a subset of it. One challenge of global interpretations is that producing global explanations may be very difficult when the underlying AI model is complex (Ribeiro et al., 2016).

2) **Local** or **prediction-level** explainability focuses on a specific decision rather than explaining the whole behaviour of an AI model. They are mostly useful when the practitioner wants to know what features or interactions cause a particular decision. It is sometimes possible to aggregate local interpretations in order to have a global interpretation of the original model. A main challenge of local interpretations is that there may be some degree of inconsistency between local explanations because a complex model may use a feature in different ways with respect to the other features (Ribeiro et al., 2016).

In a recent survey, Guidotti et al. (Guidotti et al., 2018) define the model-based interpretability as 'transparent-box design' that refers to directly designing a transparent model with the goal of solving the same problem that can be properly solved by a black-box model. On the other hand, the post-hoc interpretability is defined as 'black-box explanation' that refers to investigating how the AI system returned certain outcomes. They further divide the post-hoc explainability into three subclasses: 1) '**model explanation**' that concerns explaining the whole logic of the black-box AI system, 2) '**outcome explanation**' that refers to understanding the reasons for the decisions made for specific data records, and 3) '**model inspection**' that investigates the internal behaviour of the black-box model when the input is changed.

*4.2. Explainable artificial intelligence in the biomedical domain*

As discussed in previous sections, AI and deep learning methods have been widely utilized for building intelligent systems for biomedical and clinical applications. However, a main challenge is that powerful AI systems are usually black-box, and machine decisions are oftentimes poorly understood. This issue can be even more problematic in biomedical and healthcare domains, where critical decisions are made every day, thus, failure and mistakes can severely have bad consequences (Tjoa and Guan, 2021). Lack of transparency is one of the main barriers to widespread adoption of AI systems in clinical practice (Markus et al., 2021). Medical professionals need to understand how and why an AI system makes a particular decision (Holzinger et al., 2019). Moreover, biomedical researchers who utilize AI and machine learning methods for data analysis must have the possibility to identify bias in the data and the model, use explanations for judging the plausibility of predictions and forming novel hypotheses, and identify potential failure points of the model. The possibility to understand and explain machine decisions is the foundation of retraceability (Holzinger et al., 2019).

So far, different methods have been developed to address explainability challenges for various biomedical and healthcare applications such as visualization of pleural effusion in radiographs (Irvin et al., 2019), heatmaps for analyzing fMRI data, melanoma classification of dermoscopy images, and classification of Alzheimer's disease pathologies (Tang et al., 2019; Thomas et al., 2019; Young et al., 2019), feature relevance scores for classification of insomnia (Jansen et al., 2019), saliency maps for electroencephalogram (EEG) sleep stage scoring (Vilamala et al., 2017), visual interpretability of segmentation of tumour in brain MRI (Qin et al., 2018), constructing Bayesian rule lists for stroke prediction (Letham et al., 2015), textual justification for breast mass classification systems (H. Lee et al., 2019a), mathematical explanations for analysing metabolites signals from magnetic resonance spectroscopy and estimation of brain states (Hatami et al., 2018; Haufe et al., 2014), feature importance scores for identifying the most protective and risky genotypes for autism spectrum disorders (Ghafouri-Fard et al., 2019), and counterfactual explanations for predicting the risk of diabetes (Wachter et al., 2017).

However, most previous works on explainable AI have focused on image, continuous signal, and tabular data, hence, many challenges have remained unaddressed in biomedical and clinical text processing (Moradi and Samwald, 2021b).

# References

Emily Alsentzer, John Murphy, William Boag, Wei-Hung Weng, Di Jindi, Tristan Naumann, and Matthew McDermott. 2019. Publicly Available Clinical BERT Embeddings. In *Proceedings of the 2nd Clinical Natural Language Processing Workshop*. Association for Computational Linguistics, pages 72-78. https://doi.org/10.18653/v1/W19-1909.

Amisha, Paras Malik, Monika Pathania, and Vyas Kumar Rathaur. 2019. Overview of artificial intelligence in medicine. *Journal of family medicine and primary care, 8*(7): 2328-2331. 10.4103/jfmpc.jfmpc_440_19.

William G. Baxt. 1990. Use of an Artificial Neural Network for Data Analysis in Clinical Decision-Making: The Diagnosis of Acute Coronary Occlusion. *Neural Computation, 2*(4): 480-489. 10.1162/neco.1990.2.4.480.

Iz Beltagy, Kyle Lo, and Arman Cohan. 2019. SciBERT: A Pretrained Language Model for Scientific Text. In *Proceedings of the 2019 Conference on Empirical Methods in Natural Language Processing and the 9th International Joint Conference on Natural Language Processing (EMNLP-IJCNLP)*. Association for Computational Linguistics, pages 3615-3620. https://doi.org/10.18653/v1/D19-1371.

Steven Bethard, Guergana Savova, Wei-Te Chen, Leon Derczynski, James Pustejovsky, and Marc Verhagen. 2016. SemEval-2016 Task 12: Clinical TempEval. Association for Computational Linguistics, pages 1052-1062. 10.18653/v1/S16-1165.

Jose Camacho-Collados, and Mohammad Taher Pilehvar. 2018. From word to sense embeddings: A survey on vector representations of meaning. *Journal of Artificial Intelligence Research, 63*: 743-788.

K. R. Chowdhary. 2020. Natural Language Processing *Fundamentals of Artificial Intelligence* (pp. 603-649). New Delhi: Springer India.

Gobinda G. Chowdhury. 2003. Natural language processing. *Annual Review of Information Science and Technology, 37*(1): 51-89. https://doi.org/10.1002/aris.1440370103.

Kevin Bretonnel Cohen. 2014. Chapter 6 - Biomedical Natural Language Processing and Text Mining. In I. N. Sarkar (Ed.), *Methods in Biomedical Informatics* (pp. 141-177). Oxford: Academic Press.

Jacob Devlin, Ming-Wei Chang, Kenton Lee, and Kristina Toutanova. 2018. Bert: Pre-training of deep bidirectional transformers for language understanding. *arXiv preprint arXiv:1810.04805*.

Finale Doshi-Velez, and Been Kim. 2017. Towards a rigorous science of interpretable machine learning. *arXiv preprint arXiv:1702.08608*.

Caitlin Dreisbach, Theresa A. Koleck, Philip E. Bourne, and Suzanne Bakken. 2019. A systematic review of natural language processing and text mining of symptoms from electronic patient-authored text data. *International Journal of Medical Informatics, 125*: 37-46. https://doi.org/10.1016/j.ijmedinf.2019.02.008.

Wafaa S. El-Kassas, Cherif R. Salama, Ahmed A. Rafea, and Hoda K. Mohamed. 2021. Automatic text summarization: A comprehensive survey. *Expert Systems with Applications, 165*: 113679. https://doi.org/10.1016/j.eswa.2020.113679.

Ehsan Fathi, and Babak Maleki Shoja. 2018. Chapter 9 - Deep Neural Networks for Natural Language Processing. In V. N. Gudivada & C. R. Rao (Eds.), *Handbook of Statistics* (Vol. 38, pp. 229-316): Elsevier.

Li Feng, Li Qicheng, Mei Lijun, Li ShaoChun, Rong Liu, Chen Weiye, and Wang Fen Fei. 2015. Second Order-Based Real-Time Anomaly Detection for Application Maintenance Services. In *2015 International Conference on Service Science (ICSS)*. pages 37-44. 10.1109/ICSS.2015.23.

Wilco W. M. Fleuren, and Wynand Alkema. 2015. Application of text mining in the biomedical domain. *Methods, 74*: 97-106. https://doi.org/10.1016/j.ymeth.2015.01.015.

Wilco W. M. Fleuren, Stefan Verhoeven, Raoul Frijters, Bart Heupers, Jan Polman, René van Schaik, Jacob de Vlieg, and Wynand Alkema. 2011. CoPub update: CoPub 5.0 a text mining system to answer biological questions. *Nucleic Acids Research, 39*(suppl_2): W450-W454. 10.1093/nar/gkr310.

Deep Learning, Natural Language Processing, and Explainable Artificial Intelligence in the Biomedical Domain
Jean-Fred Fontaine, Adriano Barbosa-Silva, Martin Schaefer, Matthew R. Huska, Enrique M. Muro, and Miguel A. Andrade-Navarro. 2009. MedlineRanker: flexible ranking of biomedical literature. *Nucleic Acids Research,* *37*(suppl_2): W141-W146. 10.1093/nar/gkp353.

P. Fontelo, F. Liu, and M. Ackerman. 2005. askMEDLINE: A free-text, natural language query tool for MEDLINE/PubMed. *BMC Medical Informatics and Decision Making, 5.* 10.1186/1472-6947-5-5.

Soudeh Ghafouri-Fard, Mohammad Taheri, Mir Davood Omrani, Amir Daaee, Hossein Mohammad-Rahimi, and Hosein Kazazi. 2019. Application of Single-Nucleotide Polymorphisms in the Diagnosis of Autism Spectrum Disorders: A Preliminary Study with Artificial Neural Networks. *Journal of Molecular Neuroscience,* *68*(4): 515-521. 10.1007/s12031-019-01311-1.

Riccardo Guidotti, Anna Monreale, Salvatore Ruggieri, Franco Turini, Fosca Giannotti, and Dino Pedreschi. 2018. A Survey of Methods for Explaining Black Box Models. *ACM Comput. Surv., 51*(5): 1-42. https://doi.org/10.1145/3236009.

A. A. Gunn. 1976. The diagnosis of acute abdominal pain with computer analysis. *Journal of the Royal College of Surgeons of Edinburgh, 21*(3): 170-172.

Udo Hahn, and Michel Oleynik. 2020. Medical Information Extraction in the Age of Deep Learning. *Yearbook of Medical Informatics, 29*(01): 208-220.

Pavel Hamet, and Johanne Tremblay. 2017. Artificial intelligence in medicine. *Metabolism, 69*: S36-S40. https://doi.org/10.1016/j.metabol.2017.01.011.

Nima Hatami, Michaël Sdika, and Hélène Ratiney. 2018. Magnetic Resonance Spectroscopy Quantification Using Deep Learning. Springer International Publishing, pages 467-475.

Stefan Haufe, Frank Meinecke, Kai Görgen, Sven Dähne, John-Dylan Haynes, Benjamin Blankertz, and Felix Bießmann. 2014. On the interpretation of weight vectors of linear models in multivariate neuroimaging. *NeuroImage, 87*: 96-110. https://doi.org/10.1016/j.neuroimage.2013.10.067.

Karsten Hokamp, and Kenneth H. Wolfe. 2004. PubCrawler: keeping up comfortably with PubMed and GenBank. *Nucleic Acids Research, 32*(suppl_2): W16-W19. 10.1093/nar/gkh453.

Andreas Holzinger, Chris Biemann, Constantinos S Pattichis, and Douglas B Kell. 2017. What do we need to build explainable AI systems for the medical domain? *arXiv preprint arXiv:1712.09923*.

Andreas Holzinger, Georg Langs, Helmut Denk, Kurt Zatloukal, and Heimo Müller. 2019. Causability and explainability of artificial intelligence in medicine. *WIREs Data Mining and Knowledge Discovery, 9*(4): e1312. https://doi.org/10.1002/widm.1312.

J. J. Hopfield. 1988. Artificial neural networks. *IEEE Circuits and Devices Magazine, 4*(5): 3-10. 10.1109/101.8118.

Ke-Chun Huang, I. Jen Chiang, Furen Xiao, Chun-Chih Liao, Charles Chih-Ho Liu, and Jau-Min Wong. 2013. PICO element detection in medical text without metadata: Are first sentences enough? *Journal of Biomedical Informatics, 46*(5): 940-946. https://doi.org/10.1016/j.jbi.2013.07.009.

Kexin Huang, Abhishek Singh, Sitong Chen, Edward Moseley, Chih-Ying Deng, Naomi George, and Charolotta Lindvall. 2020. Clinical XLNet: Modeling Sequential Clinical Notes and Predicting Prolonged Mechanical Ventilation. In *Proceedings of the 3rd Clinical Natural Language Processing Workshop*. Association for Computational Linguistics, pages 94-100. https://doi.org/10.18653/v1/2020.clinicalnlp-1.11.

Jeremy Irvin, Pranav Rajpurkar, Michael Ko, Yifan Yu, Silviana Ciurea-Ilcus, Chris Chute, Henrik Marklund, Behzad Haghgoo, Robyn Ball, Katie Shpanskaya, Jayne Seekins, David A. Mong, Safwan S. Halabi, Jesse K. Sandberg, Ricky Jones, David B. Larson, Curtis P. Langlotz, Bhavik N. Patel, Matthew P. Lungren, and Andrew Y. Ng. 2019. CheXpert: A Large Chest Radiograph Dataset with Uncertainty Labels and Expert Comparison. *Proceedings of the AAAI Conference on Artificial Intelligence, 33*(01): 590-597. 10.1609/aaai.v33i01.3301590.

Christoph Jansen, Thomas Penzel, Stephan Hodel, Stefanie Breuer, Martin Spott, and Dagmar Krefting. 2019. Network physiology in insomnia patients: Assessment of relevant changes in network topology with interpretable machine learning models. *Chaos: An Interdisciplinary Journal of Nonlinear Science, 29*(12): 123129. 10.1063/1.5128003.

Xiaonan Ji, Alan Ritter, and Po-Yin Yen. 2017. Using ontology-based semantic similarity to facilitate the article screening process for systematic reviews. *Journal of Biomedical Informatics, 69*: 33-42. https://doi.org/10.1016/j.jbi.2017.03.007.

Qiao Jin, Bhuwan Dhingra, William Cohen, and Xinghua Lu. 2019. Probing Biomedical Embeddings from Language Models. In *Proceedings of the 3rd Workshop on Evaluating Vector Space Representations for NLP*. Association for Computational Linguistics, pages 82-89. https://doi.org/10.18653/v1/W19-2011.

Abdulrahman Khalifa, and Stéphane Meystre. 2015. Adapting existing natural language processing resources for cardiovascular risk factors identification in clinical notes. *Journal of Biomedical Informatics, 58*: S128-S132. https://doi.org/10.1016/j.jbi.2015.08.002.

Aditya Khamparia, and Karan Mehtab Singh. 2019. A systematic review on deep learning architectures and applications. *Expert Systems, 36*(3): e12400. https://doi.org/10.1111/exsy.12400.

Faiza Khan Khattak, Serena Jeblee, Chloé Pou-Prom, Mohamed Abdalla, Christopher Meaney, and Frank Rudzicz. 2019. A survey of word embeddings for clinical text. *Journal of Biomedical Informatics, 100*: 100057. https://doi.org/10.1016/j.yjbinx.2019.100057.

Vishesh Kumar, Katherine Liao, Su-Chun Cheng, Sheng Yu, Uri Kartoun, Ari Brettman, Vivian Gainer, Andrew Cagan, Shawn Murphy, Guergana Savova, Pei Chen, Peter Szolovits, Zongqi Xia, Elizabeth Karlson, Robert Plenge, Ashwin Ananthakrishnan, Susanne Churchill, Tianxi Cai, Isaac Kohane, and Stanley Shaw. 2014. Natural language processing improves phenotypic accuracy in an electronic medical record cohort of type 2 diabetes and cardiovascular disease. *Journal of the American College of Cardiology, 63*(12, Supplement): A1359. https://doi.org/10.1016/S0735-1097(14)61359-0.

Himabindu Lakkaraju, Ece Kamar, Rich Caruana, and Jure Leskovec. 2017. Interpretable & explorable approximations of black box models. *arXiv preprint arXiv:1707.01154*.

Yann LeCun, Yoshua Bengio, and Geoffrey Hinton. 2015. Deep learning. *Nature, 521*(7553): 436-444. https://doi.org/10.1038/nature14539.

Hyebin Lee, Seong Tae Kim, and Yong Man Ro. 2019a. Generation of Multimodal Justification Using Visual Word Constraint Model for Explainable Computer-Aided Diagnosis. Springer International Publishing, pages 21-29.

Jinhyuk Lee, Wonjin Yoon, Sungdong Kim, Donghyeon Kim, Sunkyu Kim, Chan Ho So, and Jaewoo Kang. 2019b. BioBERT: a pre-trained biomedical language representation model for biomedical text mining. *Bioinformatics, 36*(4): 1234-1240. https://doi.org/10.1093/bioinformatics/btz682.

Benjamin Letham, Cynthia Rudin, Tyler H. McCormick, and David Madigan. 2015. Interpretable classifiers using rules and Bayesian analysis: Building a better stroke prediction model. *The Annals of Applied Statistics, 9*(3): 1350-1371, 1322.

C. Li, A. Jimeno-Yepes, M. Arregui, H. Kirsch, and D. Rebholz-Schuhmann. 2013. PCorral - Interactive mining of protein interactions from MEDLINE. *Database, 2013*. 10.1093/database/bat030.

C. Lin, D. Dligach, T. A. Miller, S. Bethard, and G. K. Savova. 2016. Multilayered temporal modeling for the clinical domain. *Journal of the American Medical Informatics Association, 23*(2): 387-395. 10.1093/jamia/ocv113.

Zachary C Lipton. 2016. The mythos of model interpretability. *arXiv preprint arXiv:1606.03490*.

Weibo Liu, Zidong Wang, Xiaohui Liu, Nianyin Zeng, Yurong Liu, and Fuad E. Alsaadi. 2017. A survey of deep neural network architectures and their applications. *Neurocomputing, 234*: 11-26. https://doi.org/10.1016/j.neucom.2016.12.038.

Aniek F. Markus, Jan A. Kors, and Peter R. Rijnbeek. 2021. The role of explainability in creating trustworthy artificial intelligence for health care: A comprehensive survey of the terminology, design choices, and evaluation strategies. *Journal of Biomedical Informatics, 113*: 103655. https://doi.org/10.1016/j.jbi.2020.103655.

Warren S. McCulloch, and Walter Pitts. 1943. A logical calculus of the ideas immanent in nervous activity. *The bulletin of mathematical biophysics, 5*(4): 115-133. 10.1007/BF02478259.

Tomas Mikolov, Ilya Sutskever, Kai Chen, Greg S Corrado, and Jeff Dean. 2013. Distributed representations of words and phrases and their compositionality. In *Advances in neural information processing systems*. pages 3111-3119.

Seonwoo Min, Byunghan Lee, and Sungroh Yoon. 2017. Deep learning in bioinformatics. *Briefings in bioinformatics, 18*(5): 851-869. https://doi.org/10.1093/bib/bbw068.

Deep Learning, Natural Language Processing, and Explainable Artificial Intelligence in the Biomedical Domain

Yoav Mintz, and Ronit Brodie. 2019. Introduction to artificial intelligence in medicine. *Minimally Invasive Therapy & Allied Technologies, 28*(2): 73-81. 10.1080/13645706.2019.1575882.

Rashmi Mishra, Jiantao Bian, Marcelo Fiszman, Charlene R. Weir, Siddhartha Jonnalagadda, Javed Mostafa, and Guilherme Del Fiol. 2014. Text summarization in the biomedical domain: A systematic review of recent research. *Journal of Biomedical Informatics, 52*: 457-467. https://doi.org/10.1016/j.jbi.2014.06.009.

Milad Moradi. 2018. CIBS: A biomedical text summarizer using topic-based sentence clustering. *Journal of Biomedical Informatics, 88*: 53-61. https://doi.org/10.1016/j.jbi.2018.11.006.

Milad Moradi, Maedeh Dashti, and Matthias Samwald. 2020a. Summarization of biomedical articles using domain-specific word embeddings and graph ranking. *Journal of Biomedical Informatics, 107*: 103452. https://doi.org/10.1016/j.jbi.2020.103452.

Milad Moradi, Georg Dorffner, and Matthias Samwald. 2020b. Deep contextualized embeddings for quantifying the informative content in biomedical text summarization. *Computer Methods and Programs in Biomedicine, 184*: 105117. https://doi.org/10.1016/j.cmpb.2019.105117.

Milad Moradi, and Nasser Ghadiri. 2017. Quantifying the informativeness for biomedical literature summarization: An itemset mining method. *Computer Methods and Programs in Biomedicine, 146*: 77-89. https://doi.org/10.1016/j.cmpb.2017.05.011.

Milad Moradi, and Nasser Ghadiri. 2018. Different approaches for identifying important concepts in probabilistic biomedical text summarization. *Artificial Intelligence in Medicine, 84*: 101-116. https://doi.org/10.1016/j.artmed.2017.11.004.

Milad Moradi, and Matthias Samwald. 2021a. Evaluating the Robustness of Neural Language Models to Input Perturbations. Association for Computational Linguistics, pages 1558-1570.

Milad Moradi, and Matthias Samwald. 2021b. Explaining Black-Box Models for Biomedical Text Classification. *IEEE Journal of Biomedical and Health Informatics, 25*(8): 3112-3120. https://doi.org/10.1109/JBHI.2021.3056748.

Milad Moradi, and Matthias Samwald. 2021c. Post-hoc explanation of black-box classifiers using confident itemsets. *Expert Systems with Applications, 165*: 113941. https://doi.org/10.1016/j.eswa.2020.113941.

W James Murdoch, Chandan Singh, Karl Kumbier, Reza Abbasi-Asl, and Bin Yu. 2019. Interpretable machine learning: definitions, methods, and applications. *arXiv preprint arXiv:1901.04592*.

Prakash M Nadkarni, Lucila Ohno-Machado, and Wendy W Chapman. 2011. Natural language processing: an introduction. *Journal of the American Medical Informatics Association, 18*(5): 544-551. 10.1136/amiajnl-2011-000464.

D. W. Otter, J. R. Medina, and J. K. Kalita. 2021. A Survey of the Usages of Deep Learning for Natural Language Processing. *IEEE Transactions on Neural Networks and Learning Systems, 32*(2): 604-624. 10.1109/TNNLS.2020.2979670.

Jeffrey Pennington, Richard Socher, and Christopher D Manning. 2014. Glove: Global vectors for word representation. In *Proceedings of the 2014 conference on empirical methods in natural language processing (EMNLP)*. pages 1532-1543.

Matthew Peters, Mark Neumann, Mohit Iyyer, Matt Gardner, Christopher Clark, Kenton Lee, and Luke Zettlemoyer. 2018. Deep Contextualized Word Representations. In *Proceedings of the 2018 Conference of the North American Chapter of the Association for Computational Linguistics: Human Language Technologies*. Association for Computational Linguistics, pages 2227-2237. https://doi.org/10.18653/v1/N18-1202.

Laura Plaza. 2014. Comparing different knowledge sources for the automatic summarization of biomedical literature. *Journal of Biomedical Informatics, 52*: 319-328. https://doi.org/10.1016/j.jbi.2014.07.014.

Laura Plaza, Alberto Díaz, and Pablo Gervás. 2011. A semantic graph-based approach to biomedical summarisation. *Artificial Intelligence in Medicine, 53*(1): 1-14. https://doi.org/10.1016/j.artmed.2011.06.005.

M. V. Plikus, Z. Zhang, and C. M. Chuong. 2006. PubFocus: Semantic MEDLINE/PubMed citations analytics through integration of controlled biomedical dictionaries and ranking algorithm. *BMC Bioinformatics, 7*. 10.1186/1471-2105-7-424.

Lori L Popejoy, Mohammed A Khalilia, Mihail Popescu, Colleen Galambos, Vanessa Lyons, Marilyn Rantz, Lanis Hicks, and Frank Stetzer. 2014. Quantifying care coordination using natural language processing and domain-specific ontology. *Journal of the American Medical Informatics Association, 22*(e1): e93-e103. 10.1136/amiajnl-2014-002702.

Sameer Pradhan, Noemie Elhadad, Brett R South, David Martinez, Lee M Christensen, Amy Vogel, Hanna Suominen, Wendy W Chapman, and Guergana K Savova. 2013. Task 1: ShARe/CLEF eHealth Evaluation Lab 2013. In *CLEF (Working Notes)*. pages 212-231.

Alun Preece, Dan Harborne, Dave Braines, Richard Tomsett, and Supriyo Chakraborty. 2018. Stakeholders in explainable AI. *arXiv preprint arXiv:1810.00184*.

Yao Qin, Konstantinos Kamnitsas, Siddharth Ancha, Jay Nanavati, Garrison Cottrell, Antonio Criminisi, and Aditya Nori. 2018. Autofocus Layer for Semantic Segmentation. Springer International Publishing, pages 603-611.

Alec Radford, Jeffrey Wu, Rewon Child, David Luan, Dario Amodei, and Ilya Sutskever. 2019. Language models are unsupervised multitask learners. *OpenAI blog, 1*(8): 9.

K. Raja, S. Subramani, and J. Natarajan. 2013. PPInterFinder - A mining tool for extracting causal relations on human proteins from literature. *Database, 2013.* 10.1093/database/bas052.

A. N. Ramesh, C. Kambhampati, J. R. T. Monson, and P. J. Drew. 2004. Artificial intelligence in medicine. *Annals of the Royal College of Surgeons of England, 86*(5): 334-338. 10.1308/147870804290.

Trisevgeni Rapakoulia, Konstantinos Theofilatos, Dimitrios Kleftogiannis, Spiros Likothanasis, Athanasios Tsakalidis, and Seferina Mavroudi. 2014. EnsembleGASVR: a novel ensemble method for classifying missense single nucleotide polymorphisms. *Bioinformatics, 30*(16): 2324-2333. 10.1093/bioinformatics/btu297.

D. Ravì, C. Wong, F. Deligianni, M. Berthelot, J. Andreu-Perez, B. Lo, and G. Yang. 2017. Deep Learning for Health Informatics. *IEEE Journal of Biomedical and Health Informatics, 21*(1): 4-21. https://doi.org/10.1109/JBHI.2016.2636665.

Lawrence H. Reeve, Hyoil Han, and Ari D. Brooks. 2007. The use of domain-specific concepts in biomedical text summarization. *Information Processing & Management, 43*(6): 1765-1776. https://doi.org/10.1016/j.ipm.2007.01.026.

Marco Tulio Ribeiro, Sameer Singh, and Carlos Guestrin. 2016. Model-agnostic interpretability of machine learning. *arXiv preprint arXiv:1606.05386*.

Stuart J. Russell, and Peter Norvig. 2009. *Artificial Intelligence: A Modern Approach (3rd ed.)*. Upper Saddle River, New Jersey: Prentice Hall.

Hojjat Salmasian, Daniel E Freedberg, Julian A Abrams, and Carol Friedman. 2013. An automated tool for detecting medication overuse based on the electronic health records. *Pharmacoepidemiology and Drug Safety, 22*(2): 183-189. https://doi.org/10.1002/pds.3387.

Amber Stubbs, Christopher Kotfila, Hua Xu, and Özlem Uzuner. 2015. Identifying risk factors for heart disease over time: Overview of 2014 i2b2/UTHealth shared task Track 2. *Journal of Biomedical Informatics, 58*: S67-S77. https://doi.org/10.1016/j.jbi.2015.07.001.

Ziqi Tang, Kangway V. Chuang, Charles DeCarli, Lee-Way Jin, Laurel Beckett, Michael J. Keiser, and Brittany N. Dugger. 2019. Interpretable classification of Alzheimer's disease pathologies with a convolutional neural network pipeline. *Nature Communications, 10*(1): 2173. 10.1038/s41467-019-10212-1.

Noha S. Tawfik, and Marco R. Spruit. 2020. Evaluating sentence representations for biomedical text: Methods and experimental results. *Journal of Biomedical Informatics, 104*: 103396. https://doi.org/10.1016/j.jbi.2020.103396.

Konstantinos Theofilatos, Niki Pavlopoulou, Christoforos Papasavvas, Spiros Likothanassis, Christos Dimitrakopoulos, Efstratios Georgopoulos, Charalampos Moschopoulos, and Seferina Mavroudi. 2015. Predicting protein complexes from weighted protein–protein interaction graphs with a novel unsupervised methodology: Evolutionary enhanced Markov clustering. *Artificial Intelligence in Medicine, 63*(3): 181-189. https://doi.org/10.1016/j.artmed.2014.12.012.

Armin W. Thomas, Hauke R. Heekeren, Klaus-Robert Müller, and Wojciech Samek. 2019. Analyzing Neuroimaging Data Through Recurrent Deep Learning Models. *Frontiers in Neuroscience, 13*(1321). 10.3389/fnins.2019.01321.

E. Tjoa, and C. Guan. 2021. A Survey on Explainable Artificial Intelligence (XAI): Toward Medical XAI. *IEEE Transactions on Neural Networks and Learning Systems, 32*(11): 4793-4813. 10.1109/TNNLS.2020.3027314.

Alan Turing. 1950. Computing Machinery and Intelligence. *Mind, LIX*(236): 433-460. 10.1093/mind/LIX.236.433.

Özlem Uzuner. 2009. Recognizing Obesity and Comorbidities in Sparse Data. *Journal of the American Medical Informatics Association, 16*(4): 561-570. https://doi.org/10.1197/jamia.M3115.

Özlem Uzuner, Yuan Luo, and Peter Szolovits. 2007. Evaluating the State-of-the-Art in Automatic De-identification. *Journal of the American Medical Informatics Association, 14*(5): 550-563. https://doi.org/10.1197/jamia.M2444.

Özlem Uzuner, Imre Solti, and Eithon Cadag. 2010. Extracting medication information from clinical text. *Journal of the American Medical Informatics Association, 17*(5): 514-518. 10.1136/jamia.2010.003947.

Özlem Uzuner, Brett R South, Shuying Shen, and Scott L DuVall. 2011. 2010 i2b2/VA challenge on concepts, assertions, and relations in clinical text. *Journal of the American Medical Informatics Association, 18*(5): 552-556. 10.1136/amiajnl-2011-000203.

Ashish Vaswani, Noam Shazeer, Niki Parmar, Jakob Uszkoreit, Llion Jones, Aidan N Gomez, Łukasz Kaiser, and Illia Polosukhin. 2017. Attention is all you need. In *Advances in neural information processing systems*. pages 5998-6008.

A. Vilamala, K. H. Madsen, and L. K. Hansen. 2017. Deep convolutional neural networks for interpretable analysis of EEG sleep stage scoring. In *2017 IEEE 27th International Workshop on Machine Learning for Signal Processing (MLSP)*. pages 1-6. 10.1109/MLSP.2017.8168133.

Sandra Wachter, Brent Mittelstadt, and Chris Russell. 2017. Counterfactual explanations without opening the black box: Automated decisions and the GDPR. *Harv. JL & Tech., 31*: 841.

Yanshan Wang, Sijia Liu, Naveed Afzal, Majid Rastegar-Mojarad, Liwei Wang, Feichen Shen, Paul Kingsbury, and Hongfang Liu. 2018a. A comparison of word embeddings for the biomedical natural language processing. *Journal of Biomedical Informatics, 87*: 12-20. https://doi.org/10.1016/j.jbi.2018.09.008.

Yanshan Wang, Liwei Wang, Majid Rastegar-Mojarad, Sungrim Moon, Feichen Shen, Naveed Afzal, Sijia Liu, Yuqun Zeng, Saeed Mehrabi, Sunghwan Sohn, and Hongfang Liu. 2018b. Clinical information extraction applications: A literature review. *Journal of Biomedical Informatics, 77*: 34-49. https://doi.org/10.1016/j.jbi.2017.11.011.

Weijian Xuan, Manhong Dai, Barbara Mirel, Justin Wilson, Brian Athey, Stanley J. Watson, and Fan Meng. An Active Visual Search Interface for MEDLINE *Computational Systems Bioinformatics* (pp. 359-369).

Daniel L. K. Yamins, and James J. DiCarlo. 2016. Using goal-driven deep learning models to understand sensory cortex. *Nature Neuroscience, 19*(3): 356-365. 10.1038/nn.4244.

Wen-wai Yim, Meliha Yetisgen, William P. Harris, and Sharon W. Kwan. 2016. Natural Language Processing in Oncology: A Review. *JAMA Oncology, 2*(6): 797-804. 10.1001/jamaoncol.2016.0213.

Kyle Young, Gareth Booth, Becks Simpson, Reuben Dutton, and Sally Shrapnel. 2019. Deep Neural Network or Dermatologist? Springer International Publishing, pages 48-55.

Han Zhang, Marcelo Fiszman, Dongwook Shin, Christopher M. Miller, Graciela Rosemblat, and Thomas C. Rindflesch. 2011. Degree centrality for semantic abstraction summarization of therapeutic studies. *Journal of Biomedical Informatics, 44*(5): 830-838. https://doi.org/10.1016/j.jbi.2011.05.001.